%% file: aamas_23_workshop.tex
\documentclass[sigconf]{aamas} 

\usepackage{balance} 
\input{packages}

\input{cmd}

\setcopyright{none}
\acmConference[ALA '23]{Proc.\@ of the Adaptive and Learning Agents Workshop (ALA 2023)}
{May 29-30, 2023}{London, UK, \url{https://alaworkshop2023.github.io/}}{Cruz, Hayes, Wang, Yates (eds.)}
\copyrightyear{2023}
\acmYear{2023}
\acmDOI{}
\acmPrice{}
\acmISBN{}
\title[Know your Enemy]{Know your Enemy: Investigating Monte-Carlo Tree Search\\ with Opponent Models in Pommerman}

\author{Jannis Weil\textsuperscript{$\ast$}\,\orcidlink{0000-0001-5439-9131}}
\affiliation{
  \institution{Technical University of Darmstadt}
  \city{Darmstadt}
  \country{Germany}}
\email{jannis.weil@tu-darmstadt.de}

\author{Johannes Czech\textsuperscript{$\ast$}\,\orcidlink{0000-0002-9568-8965}}
\affiliation{
  \institution{Technical University of Darmstadt}
  \city{Darmstadt}
  \country{Germany}}
\email{johannes.czech@cs.tu-darmstadt.de}

\author{Tobias Meuser\,\orcidlink{0000-0002-2008-5932}}
\affiliation{
  \institution{Technical University of Darmstadt}
  \city{Darmstadt}
  \country{Germany}}

\author{Kristian Kersting\,\orcidlink{0000-0002-2873-9152}}
\affiliation{
  \institution{Technical University of Darmstadt, hessian.ai, DFKI}
  \city{Darmstadt}
  \country{Germany}}

\begin{abstract}
\input{abstract}
\end{abstract}


\newcommand{\BibTeX}{\rm B\kern-.05em{\sc i\kern-.025em b}\kern-.08em\TeX}

\begin{document}


\pagestyle{fancy}
\fancyhead{}


\maketitle 


\input{acro}

\input{content}



\clearpage
\begin{acks}
\input{acknowledgments}
\end{acks}



\bibliographystyle{ACM-Reference-Format} 
\bibliography{references}


\end{document}

%% file: packages.tex
\usepackage[nolist]{acronym}
\usepackage{xspace}
\usepackage{multirow}
\usepackage{comment}
\usepackage{mathtools}
\usepackage{orcidlink}
\usepackage{bbm}
\usepackage{bigdelim}
\usepackage{tikz}
\usetikzlibrary{positioning}

\usepackage[lined,linesnumbered,noend]{algorithm2e}

%% file: cmd.tex
\DeclarePairedDelimiter\abs{\lvert}{\rvert}%

\newcommand{\agentsimplep}{$\texttt{SimplePy}$\xspace}
\newcommand{\agentsimplec}{$\texttt{SimpleCpp}$\xspace}
\newcommand{\agentrawnet}{\texttt{RawNet}\xspace}
\newcommand{\agentmctso}{$\texttt{SP-MCTS}$\xspace}
\newcommand{\agentmctst}{$\texttt{TP-MCTS}$\xspace}
\newcommand{\agentmctsosl}{$\texttt{SP-MCTS}$\xspace$\texttt{SL}$\xspace}
\newcommand{\agentmctstsl}{$\texttt{TP-MCTS}$\xspace$\texttt{SL}$\xspace}

\newcommand{\sltorl}{SRL\xspace}

\RestyleAlgo{ruled}
\SetAlgoLined\DontPrintSemicolon
\SetFuncSty{textsc}
\SetKwProg{Fn}{Function}{:}{}

%% file: abstract.tex
In combination with \acl{rl}, \acl{mcts} has shown to outperform human grandmasters in games such as Chess, Shogi and Go with little to no prior domain knowledge.
However, most classical use cases only feature up to two players. 
Scaling the search to an arbitrary number of players presents a computational challenge, especially if decisions have to be planned over a longer time horizon.
In this work, we investigate techniques that transform general-sum multiplayer games into single-player and two-player games that consider other agents to act according to given opponent models.
For our evaluation, we focus on the challenging Pommerman environment which involves partial observability, a long time horizon and sparse rewards.
In combination with our search methods, we investigate the phenomena of opponent modeling using heuristics and self-play.
Overall, we demonstrate the effectiveness of our multiplayer search variants both in a supervised learning and reinforcement learning setting.

%% file: acro.tex
\begin{acronym}[MCTS]
\acro{mcts}[MCTS]{Monte-Carlo Tree Search}
\acro{rl}[RL]{Reinforcement Learning}
\acro{sl}[SL]{Supervised Learning}
\acro{marl}[MARL]{Multi-Agent Reinforcement Learning}
\acro{lstm}[LSTM]{Long Short-Term Memory}
\acro{gru}[GRU]{Gated Recurrent Unit}
\acro{ffa}[FFA]{Free For All}
\acro{a2c}[A2C]{Advantage Actor-Critic}
\acro{a3c}[A3C]{Asynchronous Advantage Actor-Critic}
\acro{ppo}[PPO]{Proximal Policy Activation}
\acro{impala}[IMPALA]{Importance Weighted Actor-Learner Architecture}
\acro{brs}[BRS]{Best Reply Search}
\acro{puct}[PUCT]{Predictor Upper Confidence Bounds}
\end{acronym}

%% file: content.tex
\newcommand{\customfootnotetext}[2]{{
  \renewcommand{\thefootnote}{#1}
  \footnotetext[0]{#2}}}

\customfootnotetext{$\ast$}{These authors contributed equally to this work.}

\section{Introduction}

\ac{mcts} is widely know as a powerful search algorithm for both, single player environments such as Atari, and two player zero-sum games like the game of Go \cite{schrittwieser2020mastering}.
In environments with more players, \ac{mcts} is usually combined with domain-specific heuristics to make search feasible \cite{swiechowski2022mctsOverview}.

Multiplayer games with more than two players introduce new challenges to search-based methods.
In particular, the size of the search tree explodes when move combinations for multiple players have to be considered \cite{luckhart1986maxn}.
This leads to an exponentially increasing computational complexity of the search, depending on the number of players.
Combinations with \ac{mcts} \cite{nijssen2013phdMultiplayerMCTS} lead to improvements across multiple domains, but perform poorly under limited resources due to a shallow search depth \cite{baier2020guidingMultiplayerMCTS}.
Multiplayer search methods like Parandoid search \cite{sturtevant2000paranoidSearch}, \acl{brs} \cite{schadd2011bestReplySearch} and recent extensions \cite{baier2020opponentPruningParanoid} improve the search depth by reducing the time spent to simulate opponents and expectedly suboptimal moves.
Additionally, the search can be guided with value estimates \cite{szita2009monte, nijssen2011playout} and learned value functions \cite{petosa2019multiplayer, ozair21quantizedModels}.

With this work, we focus on the combinatorial aspect of multiplayer games and investigate learning-based \ac{mcts} variants that effectively reduce the search space to single- and two-player games.
Other players act according to given opponent models, hence drastically reducing the branching factor.
We give insights regarding the applicability in a \ac{rl} setting and evaluate our approach in the multiplayer game Pommerman \cite{resnick2018pommerman}.

We provide the following contributions:

\begin{itemize}
    \item Based on learning-based \ac{mcts}, we propose techniques to reduce the search space by transforming competitive \mbox{n-player} games to single-player and two-player games.
    \item We compare learning from demonstrations, reinforcement learning, and different opponent models in terms of resulting performance and behavior.
    \item We show that the proposed agent achieves a proficient level of play in the \ac{ffa} Pommerman environment.
\end{itemize}

\noindent Our code is available at \url{https://github.com/jw3il/PommerLearn}.
We begin by introducing our approach in a general setting.
Next, we go over our experiments in the Pommerlearn environment for both, learning from demonstration data, as well as learning in a reinforcement learning setting.
Afterwards we discuss the results and the limitations of our approaches.
At last, we present related work and conclude with an outlook for potential future work.

\begin{figure*}
    \centering
    \includegraphics[width=0.99\linewidth]{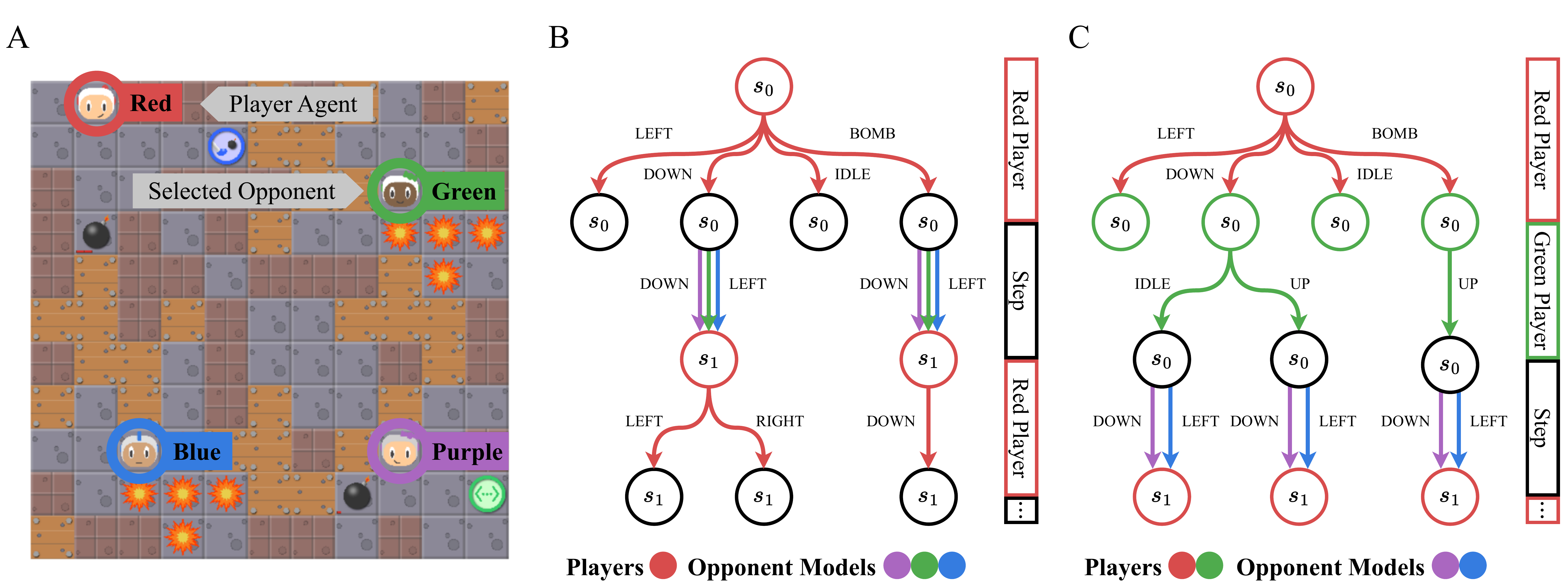}
    \caption{Exemplary search graphs of Single-Player Search (B) and Two-Player Search (C) next to a Pommerman board (A). The Single-Player Search allows a deeper search with fully heuristic-based play for all opponents, whereas the Two-Player Search allows a full exploration of a selected opponent at each step with the downside of achieving a lower search depth.}
    \Description{Figure containing three subfigures A, B, and C. Subfigure A shows a Pommerman board with four players labeled red, green, blue and purple. The red player is marked as the player agent and the green agent as the selected opponent. Subfigure B shows an exemplary search graph for the Single-Player Search approach. The deepest path is expanded over the actions of the red player, the opponent models of the purple, green, blue players and then again the red player. Subfigure C shows an exemplary search graph for the Two-Player Search approach. The deepest path is expanded over the actions of the red player, the green player, and then the opponent models of the purple and blue players. The Single-Player Search allows a deeper search with fully heuristic-based play for all opponents, whereas the Two-Player Search allows a full exploration of a selected opponent at each step with the downside of achieving a lower search depth.}
    \label{fig:search-graph}
\end{figure*}

\section{Approach}
When a model of the environment is available, leveraging this knowledge with model-based algorithms comes with several advantages.
Our work builds upon \ac{mcts}, a general search method that aims to find a sequence of moves leading a player to the expectantly most advantageous states, i.e.\ states in which the they can win the game.
This requires a model of the environment and a way to evaluate states, which could both be provided or learned.
By expanding potential future states, the search can make use of additional game knowledge to correct decisions where a suboptimal agent alone would fail.
This allows to filter potential dangers instead of having to face them.
Additionally, search methods usually return a principal variation, which is the sequence of future moves that are considered best under the current knowledge. 
By iterating over this sequence, it is possible to give a more throughout explanation of the planned behavior of the agent.

Search methods require a strategy for selecting and expanding nodes in a search tree.
In our case, this is provided by a neural network model that predicts value and policy distributions for each state, as we will detail later.
The search focuses on potential future states that appear to be promising, while also exploring other paths.
We use a variant of the \ac{puct} algorithm~\cite{rosin2011multi} to select and expand new nodes.
In particular, we refer to the algorithm adjusted by \citet{silver2017mastering}:
\begin{align}
    \label{eq:node_selection}
    a_t = \text{argmax}_a \left(\text{Q}(s_t,a) + U(s_t, a)\right) \; ,\\ \text{where} \quad
    U(s_t,a) = c_{\text{puct}} P(s_t,a) \frac{\sqrt{\sum_b N(s_t,b)}}{1 + N(s_t,a)}\;.
\end{align}
Here, $a_t$ refers to the selected action at time step $t$, and $\text{Q}(s_t,a)$ is the action value for action $a$ of state $s_t$. The action values of a node are updated by calculating a simple moving average of all backpropagated value estimates of its subtree.
The term $U(s_t, a)$ describes the utility function for a particular action. It is given by the product of its policy estimate $P(s_t, a)$ and the total number of visits of its parent divided by the number of visits $N(s_t, a)$ of the selected action.
This prioritizes actions that have a higher policy estimate or were chosen less frequently.
The denominator ${1 + N(s_t,a)}$ is used to avoid division by zero, and so that the nodes do not have to be fully expanded over each action. The scalar $c_\text{puct}$ is a weighting parameter which controls the amount of exploration compared to the greedy action selection of choosing the highest $\text{Q}$-value.

Extensions of \ac{mcts} to games with more than two players come with conceptual and practical challenges.
In particular, the search space grows exponentially with the number of players if all of them are considered in the search.

To address this issue, we propose two simple yet effective methods that reduce multiplayer games to single and two-player games.
They allow for the application of AlphaZero-like frameworks without major adjustments.
We describe the methods in the following sections and provide a visualization with Fig.~\ref{fig:search-graph}.

\subsection{Single-Player Monte-Carlo Tree Search}
A straightforward approach for simplifying the search is to transform the multi-player environment into a single-player environment by treating the opponents as part of the environment.
Instead of modifying the environment's dynamics, we simplify the search space by using deterministic opponent models for other players. 
This also builds the basis of our second approach, which we will introduce in the next section.
Instead of searching through all actions of all players, we limit the search to our player agent.
In Fig.~\ref{fig:search-graph} (B), this is the red player.
The search tree is expanded solely using the actions of this player.
To execute a step in the deterministic environment, we then gather actions for other players with their deterministic opponent models.
With all actions, the environment advances from state $s_0$ to the next state $s_1$.
Using this method, an n-player game effectively reduces to a single-player game.

The quality of the resulting policy depends on the given opponent model.
If the opponent model differs from the actual behavior of the opponents, the paths that are explored during search can get highly inaccurate and diverge from potential future trajectories in the real environment.
Convergence guarantees towards optimal behavior are lost and a higher search depth could even lead to deteriorations of the resulting policy.

Despite these unfavorable preconditions,
our hypothesis is that this approach can allow to assess the current situation in order to make good decisions with respect to immediate dangers and the near future.
The better the opponent model fits the behavior of our opponents, the better we can exploit their behavior.
As we simulate a single agent, we can perform many simulation steps that, although being inaccurate, could help to estimate the value of the available actions.
If one would use an optimal player as the opponent model, our agent would plan how to act in the worst-case scenario, irrespective of the actual behavior of the opponents.

Given an action space $\mathcal{A}$, the maximum branching factor per step is $\abs{\mathcal{A}}$ as we only expand moves of one agent. 

This search method is summarized in Alg.~\ref{algo:sp-mcts}.
The function \textsc{SelectLeafNode} corresponds to the node selection phase in \ac{mcts} and selects a leaf node by following Eq.~(\ref{eq:node_selection}), \textsc{SelectAction} then selects a new action in this leaf.
After gathering the remaining actions with the given opponent models, performing a step in the environment yields a new state.
This state is evaluated from the perspective of the player agent and the results are stored in a new node which is appended to the tree.
Finally, the value from the new node is backpropagated without depth-wise negation using \textsc{BackpropagateSP}, returning the updated tree.

\begin{algorithm}[t]
\caption{Single-player MCTS for multiplayer games. Shown is a single tree search update iteration (simulation).}
\label{algo:sp-mcts}

\SetKwFunction{FSPMCTS}{SP\_MCTS}
\Fn{\FSPMCTS{$tree$, $playerID$, $opponentIDs$}}{
     $node,\, state \gets$ \textsc{SelectLeafNode}($tree$) \;
     $action \gets$ \textsc{SelectAction}($node$) \;
      $actions[playerID] \gets action$ \;

    \For{$idx \in opponentIDs$}{
        $actions[idx] \gets$ \textsc{OpponentModel}($idx$, $state$) \;
    }
    $state' \gets$ \textsc{EnvironmentStep}($state$, $actions$) \;
    $result \gets$ \textsc{Evaluate}($state'$, $playerID$) \;
    $tree', node' \gets$ \textsc{ExpandTree}($tree$, $node$, $action$, $result$) \;
    \Return \textsc{BackpropagateSP}($tree'$, $node'$) 
}
\vspace{-0.1cm}
\end{algorithm}

\subsection{Two-Player Monte-Carlo Tree Search}

The main limitation of our single-player search is that the play behavior of our opponents remains deterministic, alternative moves are not considered, and it cannot converge to an optimal strategy during the search if there is a discrepancy between the opponent model and the actual opponent behavior.
To overcome these limitations to some extend, we propose an approach which we call two-player  search.
This approach expands our single-player search by exploring the moves of a selected opponent in each step, e.g.\ the green agent in Fig.~\ref{fig:search-graph} (C).
Instead of following the deterministic opponent model, the move nodes for this opponent can be fully expanded.
The selected opponent makes use of the same prior policy for move selection as our agent.
All remaining opponents perform actions according to their given models.
Note that the selected opponent can change across steps in the simulation.
Ideally, one would select the opponent that, when allowed to deviate from the given opponent model, results in the most reduction in our agent's estimated value.
This can be seen as an instance of \ac{brs} \cite{schadd2011bestReplySearch} where the opponent with the best reply is assumed to be known.
For simplicity, we choose the closest agent.
While \ac{brs} skips moves of opponents that are not selected, \ac{brs}\texttt{+}~ \cite{esser2013BRSplus} uses move orderings to select valid moves. 
We use opponent models to advance other opponents during the search.
In the example in Fig.~\ref{fig:search-graph} (C), our approach additionally expands the actions of the green player.
Like in vanilla \ac{mcts}, the values of the green player are negated and then backpropagated to the red player.
The other opponents are seen as a part of the environment during this step.

The two-player search expands moves for selected opponents, thus leading to a higher branching factor compared to the single-player search.
Given an action space $\mathcal{A}$, the maximum branching factor per step is now $\abs{\mathcal{A}}^2$. 
This is magnitudes smaller than the maximum branching factor of complete enumeration with $\abs{\mathcal{A}}^n$ for $n$ agents and of \ac{brs} with $(n - 1) \abs{\mathcal{A}}^2$.

The pseudocode is shown in Alg.~\ref{algo:tp-mcts}.
We alternately select a new action for the player agent and a selected opponent and expand the tree accordingly.
After selecting both actions, the opponent models fill in the remaining actions to perform a step in the environment.
Note that this uses regular backpropagation with negated values.

\subsection{Combination with Learned Models}
Based on the idea of AlphaZero \cite{silver2017mastering}, we leverage an agent model $f_\theta(o) = (\mathbf{p}, v)$ to predict move probabilities $\mathbf{p}$ and a value $v$ for a given observation $o$.
These predictions are used to guide the previously described search approaches.
In the \ac{puct} formula (see Eq.~\ref{eq:node_selection}), $P(s_t, a)$ evaluates to $\mathbf{p}$ and $v$ is used to update $Q(s_t, a)$ upon expanding non-terminal nodes.
The loss is defined as
\begin{equation}
\label{eq:loss}
    l = \alpha (z - v)^2 - (1 - \alpha) \mathbf{\pi}^\intercal \text{log}\, \mathbf{p}\;,
\end{equation}
where $z$ is the target value and $\pi$ the move probability according to the search. The total loss consists of both the value loss, that is given as a mean squared error, and the policy loss, that is formulated as a cross-entropy loss. Hyperparameter $\alpha$ weights the value loss, \citet{silver2017mastering} suggested using a low weight to reduce the chance of overfitting to the value target.
We iteratively update the model with the AdamW optimizer \cite{loshchilov2019AdamW}.

\begin{algorithm}[t]
\caption{Two-player MCTS for multiplayer games. Shown is a single tree search update iteration (simulation).}
\label{algo:tp-mcts}

\SetKwFunction{FTPMCTS}{TP\_MCTS}

\Fn{\FTPMCTS{$tree$, $playerID$, $opponentIDs$}}{
    $node,\, state \gets$ \textsc{SelectLeafNode}($tree$) \;
    $agentID \gets$ \textsc{GetActiveAgent}($node$) \;
    $action \gets$ \textsc{SelectAction}($node$) \;
    \uIf(\tcp*[f]{player node}){$agentID = playerID$}{
     $result \gets$ \textsc{Evaluate}($state$, $playerID$) \;
    }
    \Else(\tcp*[f]{opponent node and step}){
    $actions[agentID] \gets action$ \;
    $actions[playerID] \gets$ \textsc{GetLastAction}($node$) \;
    \For{$idx \in opponentIDs \setminus \{agentID\}$}{
    $actions[idx] \gets$ \textsc{OpponentModel}($idx$, $state$) \;}
    $state' \gets$ \textsc{EnvironmentStep}($state$, $actions$) \;
    $result \gets$ \textsc{Evaluate}($state'$, $agentID$) \;
    }
    $tree', node' \gets$ \textsc{ExpandTree}($tree$, $node$, $action$, $result$) \;
    \Return \textsc{Backpropagate}($tree'$, $node'$) 
}
\vspace{-0.1cm}

\end{algorithm}

\section{Experiments in Pommerman}
Pommerman \cite{resnick2018pommerman} is a multi-agent environment inspired by the video game series \emph{Bomberman}.
Up to four bomber agents move across a discrete grid-world and try to defeat their opponents by placing bombs.
In the \ac{ffa} mode, each agent plays on their own and observes the the whole board except for hidden power ups.
In the \emph{team} and \emph{radio} modes, there are two teams of two agents each. Agents can only observe their local surroundings up to a distance of 4 blocks from their current position, horizontally or vertically. 
In the radio mode, agents can additionally use a discrete communication channel and send six bits per step.
The \ac{ffa} mode has been used in a preliminary competition in 2018 \cite{resnick2018pommerman}, the teams mode at NeurIPS 2018 \cite{NeurIPS18Competition}
and the radio mode at NeurIPS 2019\footnote{\url{https://nips.cc/Conferences/2019/CompetitionTrack}.}.

The Pommerman environment is very challenging, mainly because it is a multiplayer game, its long time horizon of up to $800$ steps and partial observability.
With four players and $\abs{\mathcal{A}} = 6$ actions, exhaustively exploring the search tree for $10$ steps in order to see a newly placed bomb explode would require evaluating around $(6^4)^{10} \approx 1.34e^{31}$ states. 
Given the environment's time limit of $100$ milliseconds per move, there is a need for more efficient solutions.

\subsection{Training Setup}
We implement our approach on top of CrazyAra~\cite{czech2021improving}, an AlphaZero-like \ac{mcts} framework that includes several extensions.
Our agent model $f_\theta$ uses a \textit{RISEv2 mobile} architecture \cite{czech_learning_2020} adapted for the game Pommerman.
The input of the model is of board size $11\times 11$ with $23$ feature channels that encode the agent's observation.
Further details are provided in our repository.
The learning target $z$ is the outcome of an episode and either win~($1$), draw~($0$) or loss~($-1$).
Custom intermediate rewards and discounting are not used.

Each training iteration is performed in a supervised manner on given datasets according to the loss in Eq.~(\ref{eq:loss}) with $\alpha = 0.1$ to avoid overfitting to the value target.
We perform data augmentation to mirror and rotate all observations jointly with the targets to improve the sample efficiency.
Depending on the experiment, the datasets either originate from expert demonstrations or from samples generated by our search approaches.

The official Pommerman environment \cite{resnick2018pommerman} is implemented in Python and provides baseline agents called \texttt{SimpleAgent}.
Our approach is implemented in C\texttt{++} and makes use of a faster reimplementation of the Pommerman environment \cite{weil2021pomcpp2}.
This includes an agent called \texttt{SimpleUnbiasedAgent} that improves upon the provided C\texttt{++} \texttt{SimpleAgent} and reduces the decision bias depending on the agent's id.\footnote{We found that the C\texttt{++} \texttt{SimpleAgent} behaves differently depending on its agent id and start position, resulting in varying average win rates.}
Most of the results presented in the following sections use this reimplementation and the \ac{ffa} mode, but we conclude with preliminary results in the Python environment.

An overview of the considered agents is presented in Tab.~\ref{tab:pommer-opponents}.

\begin{table}[t!]
\centering
\small
\caption{Overview of used agents.}
\begin{tabular}{cp{5.8cm}} 
\toprule
 \textbf{Abbreviation} & \textbf{Description} \\ 
 \midrule
 \agentsimplep & \texttt{SimpleAgent} from the Python environment. \\
 \agentsimplec & \texttt{SimpleUnbiasedAgent} from pomcpp2. \\
 \agentrawnet & Chooses the action with the highest Q-value of the player model. \\
 \agentmctso & Our approach with the single-player search. \\
 \agentmctst & Our approach with the two-player search. \\
 \bottomrule
\end{tabular}
\label{tab:pommer-opponents}
\end{table}

\subsection{Learning from Demonstrations}
To investigate the effectiveness of our search approaches, we first study their combination with learning from demonstrations.
We generate a data set with one million samples of \agentsimplec agents playing \ac{ffa} games with random start conditions for up to $800$ steps.
This includes samples from the perspective of each player, i.e.~we collect four trajectories per episode.
The model is trained using our loss from Eq.~(\ref{eq:loss}) with \ac{sl}, where the target policy equals the actions chosen by the agents.
The resulting model is used as the player agent in conjunction with the search methods \agentmctso and \agentmctst.
For these experiments, we set the opponent models to \agentsimplec with random seeds. Thus, the search cannot foreshadow the exact moves that will be selected by the actual opponents, but captures their overall behavior.

\begin{figure}[t]
    \centering
    \includegraphics[width=\linewidth]{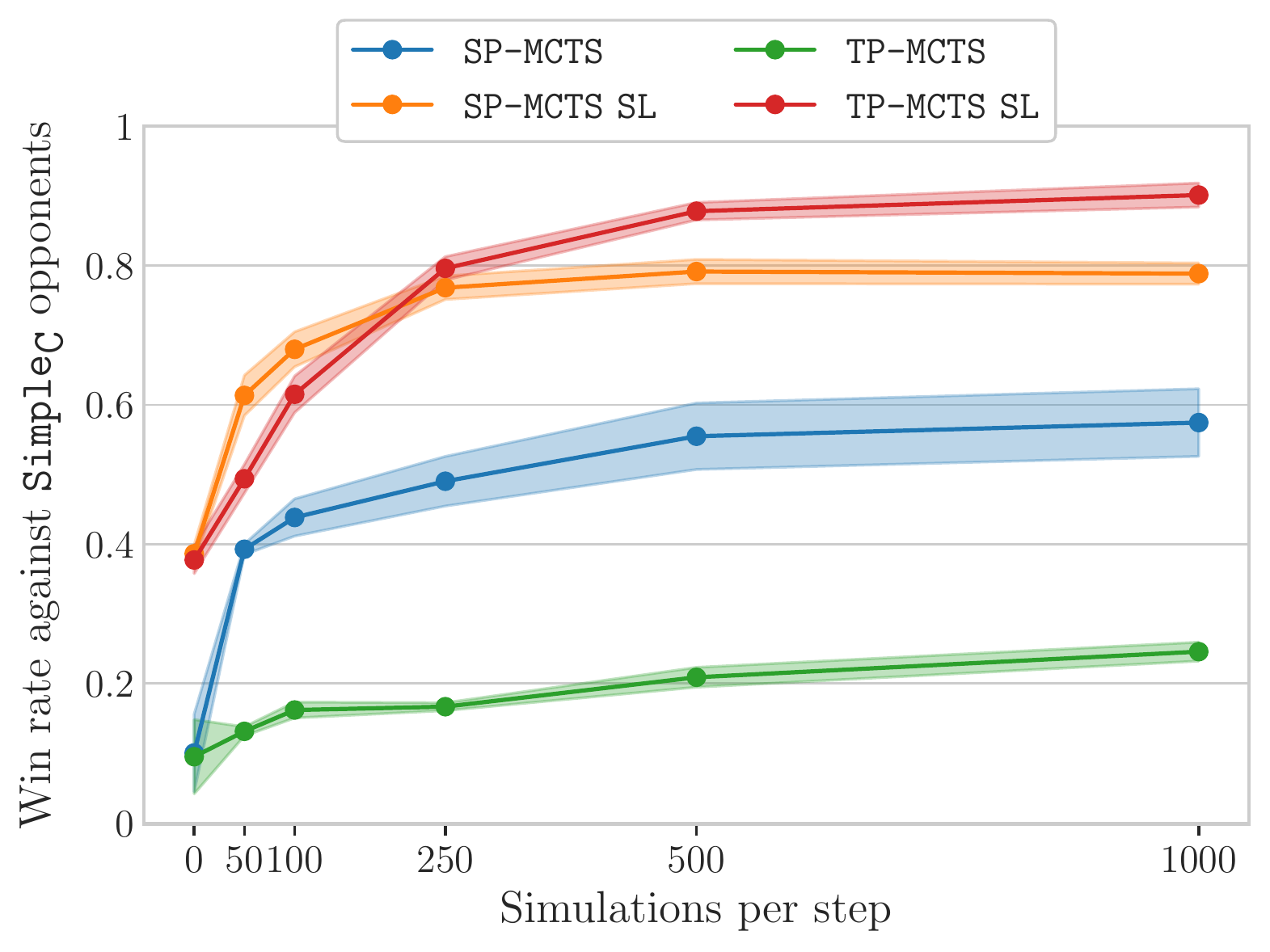}
    \caption{\looseness=-1 \texttt{TP-MCTS} \texttt{SL} outperforms \texttt{SP-MCTS} \texttt{SL} for higher num\-ber of simulations. When random initialized models are used, \agentmctso has consistently higher win rates than \agentmctst. Shown are the win rates of \agentmctso and \agentmctst against \agentsimplec in the \ac{ffa} mode. The suffix \texttt{SL} indicates that the search uses a model trained on demonstrations. The standard deviation of five runs is highlighted in the shaded area.}
    \Description{Shown is a line plot of the win rates of SP-MCTS and TP-MCTS against SimpleCpp in the FFA mode, depending on the number of search simulations per step. The plot contains the results for both, randomly initialized models and for the models trained on demonstrations. In all cases, the win rate saturates with a higher number of simulations. However, the initial win rate at 0 simulations differs. The SL models achieve a win rate of around 40 percent, while the randomly initialized models achieve a win rate around 10 percent. With a higher number of simulations and the SL models, TP-MCTS outperforms SP-MCTS. For randomly initialized models, SP-MCTS has consistently higher win rates than TP-MCTS. The standard deviation of five runs is highlighted in a shaded area, showing a small standard deviation except for the SP-MCTS runs.}
    \label{fig:search-experiments}
\end{figure}

Fig.~\ref{fig:search-experiments} shows the win rate of our approaches over $1000$ games against \agentsimplec opponents for increasing simulations per step.
For zero simulations, we use the respective \agentrawnet agent that chooses an action based on the maximum probability of the root nodes's policy distribution without any look-ahead.
The results are averaged over five models trained on the same data set with different seeds.
Note that with four agents in the \ac{ffa} mode, a win rate of $25\%$ indicates equal performance if there are no draws.
We include the results for randomly initialized models as a baseline. 

We can see in Fig.~\ref{fig:search-experiments} that for the randomly initialized models, \agentmctso highlighted in blue greatly outperforms \agentmctst highlighted in green.
This is because \agentmctst uses the given model to guide the search of the closest opponent.
Our agent tries to exploit the mistakes of its opponents.
If the opponent model is poor, the agent gets overconfident in taking bad actions and the search results do not transfer well to the real environment.

This is consistent with the comparably good results of \agentmctstsl when the expert model is used.
While \agentmctsosl outperforms \agentmctstsl for a low number of simulations, \agentmctstsl  achieves higher win rates for $250,\,500$ and $1000$ simulations per step.
The win rate for zero simulation steps of the learned model is greater than $25\%$, which indicates that its combination with action filtering already performs better than \agentsimplec.
With a high number of simulations, \agentmctstsl can reach a sufficient search depth and benefit from an increased exploration of the opponent's actions.

To summarize, the result for the model initialized with expert demonstrations are promising and we see that both search approaches greatly improve the performance of a randomly initialized model.
We now investigate whether these models can be improved by iteratively training on samples generated by the search.

\begin{figure}[tp]
    \centering
    \includegraphics[height=0.75\linewidth]{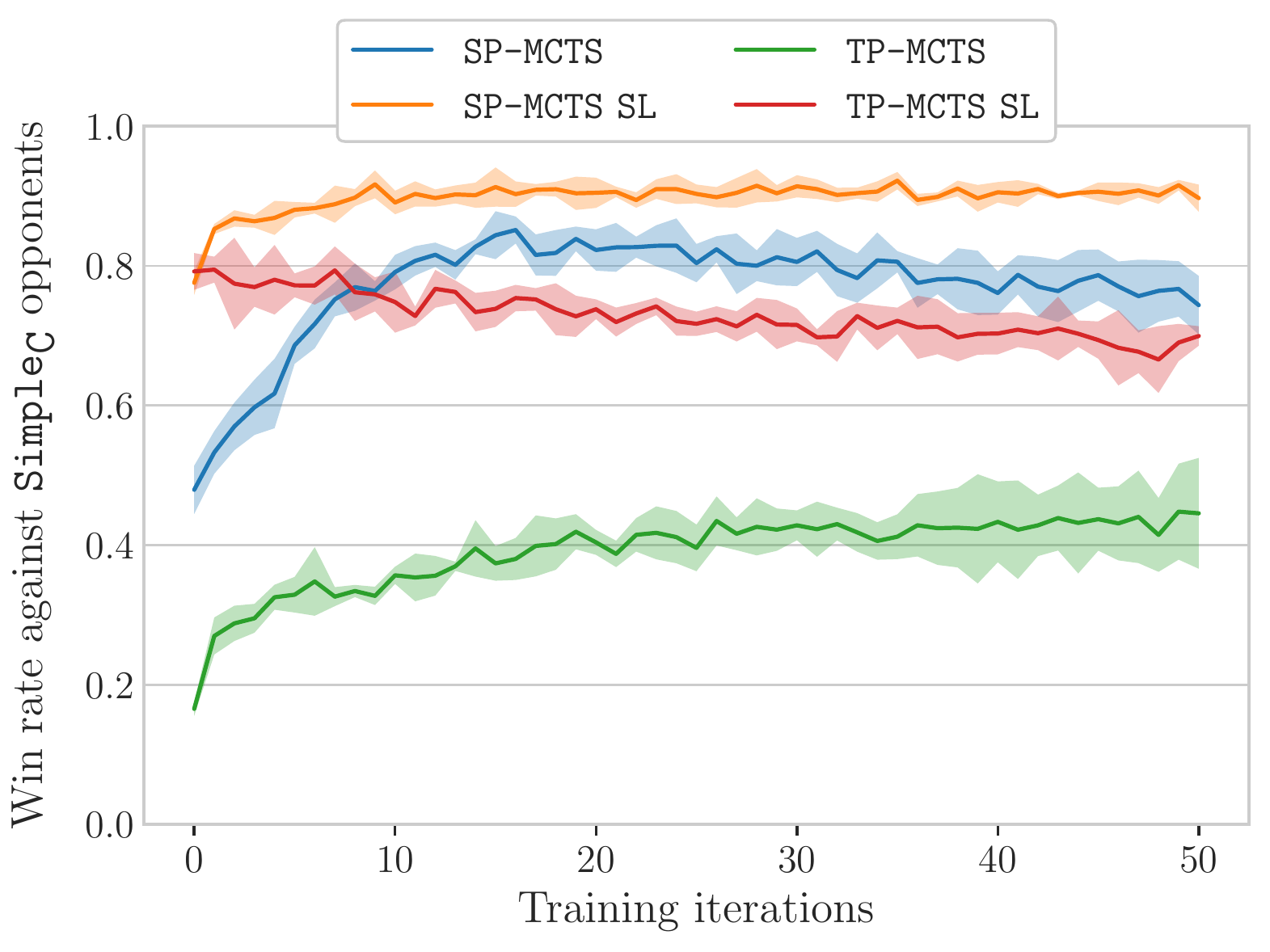}
    \caption{In reinforcement learning, both \agentmctso versions appear to perform better than \agentmctst. Shown are the win rates of RL-based \agentmctso and \agentmctst against \agentsimplec in the \ac{ffa} mode over $50$ training iterations with $250$ simulations. The configurations with suffix $\texttt{SL}$ are initialized with the model trained on demonstrations. The standard deviation of five runs is highlighted in the shaded area.}
    \Description{Shown is a line plot with the win rates of the RL-based SP-MCTS and TP-MCTS approaches against SimpleCpp in the FFA mode over 50 training iterations with 250 simulations. The plot contains the results for both, randomly initialized models and for the models trained on demonstrations. Both SP-MCTS runs and TP-MCTS with a random model improve over the initial win rates. The win rate of TP-MCTS SL decreases. The standard deviation of five runs is highlighted in a shaded area and shows that TP-MCTS has a slightly higher variance. In reinforcement learning, both SP-MCTS versions appear to perform better than TP-MCTS.}
    \label{fig:rl-experiments}
\end{figure}

\subsection{Optimization with Reinforcement Learning}
As the next step, we aim to improve the models from the previous section with \ac{rl}.
We simulate \ac{ffa} games against \agentsimplec opponents for $100\,000$ steps and use the search's results as the learning target for the policy and value functions.
The resulting model is then used in the next iteration and the process is repeated.
As before, all agent and opponent models are set to \agentsimplec.
We train our agent for $50$ iterations with $250$ simulations per step. 
This setting has been chosen as a trade-off between required simulation time and win rate based on the previous experiments.

\begin{table*}[t]
    \centering
    \caption{Win rate, tie rate, search depth, search runtime per step and environment steps for our approaches with 1000 simulations and different opponent models against \agentsimplec opponents for 1000 games.  
    All results are averaged over five models.}
    \begin{tabular}{ccccccccc}
        \toprule
        \textbf{Model} & \textbf{Method} & \textbf{Opponent Model} & \textbf{Win Rate} & \textbf{Tie Rate} & \textbf{Search Depth} & \textbf{Search Time [ms]} & \textbf{Environment Steps}\\ \midrule
        \multirow{4}{*}{SL} & \multirow{2}{*}{\agentmctso} & \agentsimplec & $0.78 \pm 0.03$ & $0.07 \pm 0.02$ & $17.91 \pm 7.60$ & $35.30 \pm 7.57$ & $188.99 \pm 3.20$\\
        & & \agentrawnet & $0.76 \pm 0.03$ & $0.10 \pm 0.02$ & $21.46 \pm 22.09$ & $266.62 \pm 138.13$ & $191.02 \pm 2.91$\\
        &\multirow{2}{*}{\agentmctst} & \agentsimplec & $\mathbf{0.91 \pm 0.01}$ & $0.06 \pm 0.01$ & $11.57 \pm 6.05$ & $36.89 \pm 7.52$ & $246.40 \pm 5.93$\\   
        & & \agentrawnet & $\mathbf{0.92 \pm 0.01}$ & $0.06 \pm 0.01$ & $12.14 \pm 7.17$ & $164.61 \pm 64.46$ & $254.79 \pm 11.26$\\\midrule
        \multirow{2}{*}{\sltorl} & \multirow{2}{*}{\agentmctso} & \agentsimplec & $\mathbf{0.94 \pm 0.01}$ & $0.02 \pm 0.01$ & $21.39 \pm 7.45$ & $37.34 \pm 6.74$ & $275.23 \pm 9.40$\\
        & & \agentrawnet & $\mathbf{0.90 \pm 0.01}$ & $0.04 \pm 0.01$ & $25.52 \pm 10.37$ & $315.74 \pm 139.75$ & $288.47 \pm 10.26$\\
        \multirow{2}{*}{RL} & \multirow{2}{*}{\agentmctso} & \agentsimplec & $0.82 \pm 0.01$ & $0.07 \pm 0.00$ & $21.98 \pm 8.65$ & $36.49 \pm 6.52$ & $331.99 \pm 7.38$\\
        & & \agentrawnet & $0.71 \pm 0.03$ & $0.08 \pm 0.01$ & $28.30 \pm 11.73$ & $309.40 \pm 140.20$ & $316.81 \pm 9.46$\\
        \bottomrule
    \end{tabular}
    \label{tab:learned-opponent-models}
\end{table*}

During our experiments, we noticed that the results highly depend on the amount of noise introduced by exploration within the search.
With the regular policy target $\pi$, the agents get stuck in local optima with low win rates after around 15 iterations. 
In the following, we focus on a configuration with a modified policy target $\pi' = 0.5\cdot \pi + 0.5\cdot \mathbbm{1}\{\pi = \max \pi \}$, as we found the corresponding results to be more insightful.
This reduces the noise introduced by the search and in turn increases the probability of choosing the best action with the highest visit count by 50\%.

The results in Fig.~\ref{fig:rl-experiments} are averaged over five runs using the models from the previous section.
We can see that the win rate of \agentmctso and \agentmctst increases, suggesting improvements of the corresponding models.
However, the win rate of \agentmctso reaches its peak at around 15 iterations and starts to slowly decline afterwards.
While the win rate of \agentmctsosl slightly increases from 80\% to 90\%, the win rate of \agentmctstsl decreases over time.
Investigating the resulting policies reveals that the agents learn to play passively with \ac{rl}, i.e.~they wait for their opponents and evade bombs when necessary.
This strategy is unexpected but expedient, as \agentsimplec opponents are suboptimal and often put themselves in unfavorable situations.

\begin{figure}[t!]
    \centering
    \includegraphics[height=0.75\linewidth]{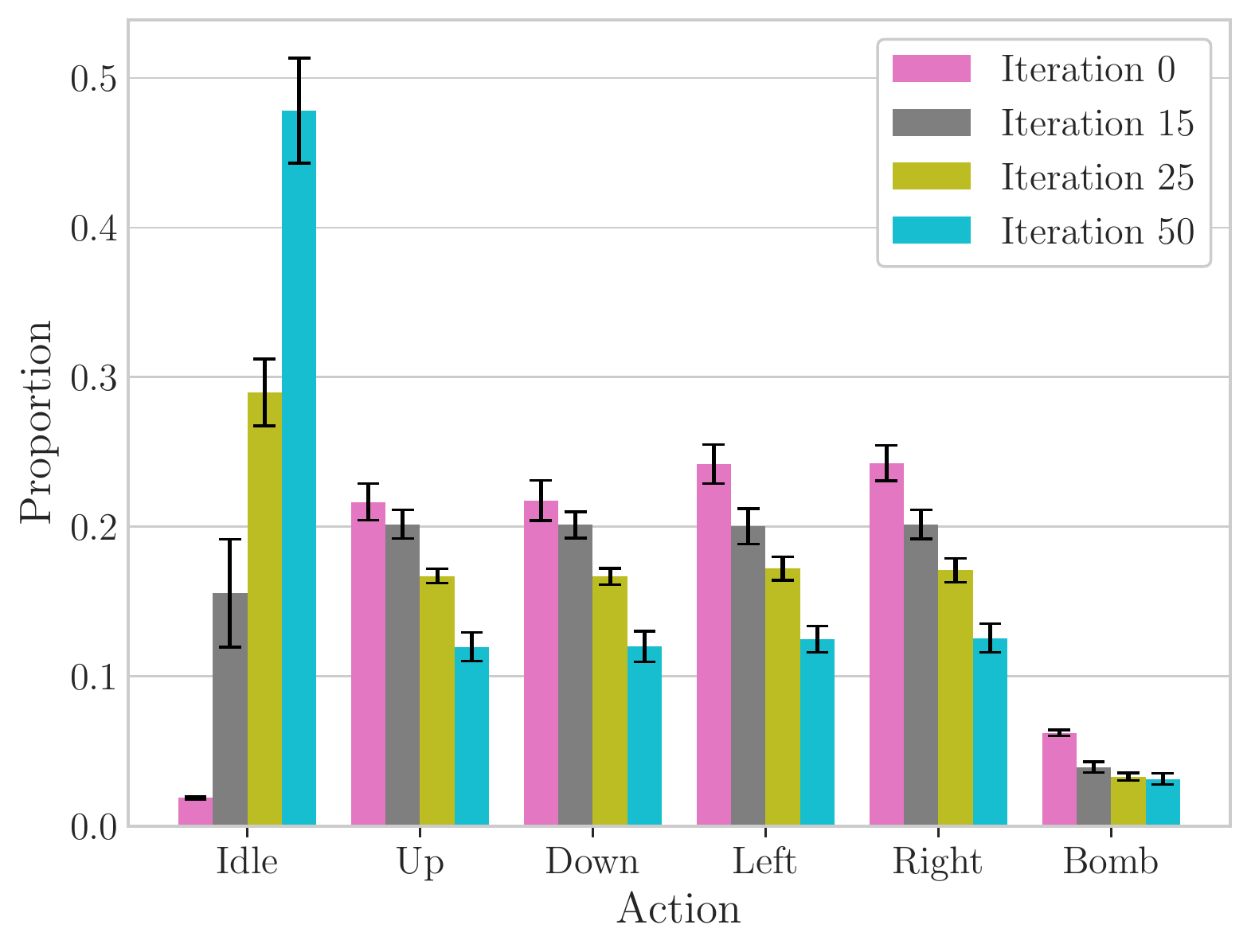}
    \caption{\agentmctsosl becomes more passive during reinforcement learning over time, reflected by the higher use of idle actions. Shown is the action distribution of \agentmctso SL during training with \ac{rl} at iterations 0, 15, 25 and 50. The standard deviation of five runs is highlighted by the error bars.}
    \Description{Shown is a bar plot of the action distribution of SP-MCTS SL during training with RL at iterations 0, 15, 25 and 50. The standard deviation of five runs is highlighted by error bars. They show a comparably higher standard deviation for idle actions. One can see that SP-MCTS becomes more passive during reinforcement learning over time, reflected by a higher use of idle actions after more training iterations.}
    \label{fig:rl-experiments-actions}
\end{figure}

This is particularly visible in the \agentmctsosl configuration with a win rate of around 90\% after training with \ac{rl}.
We show our agent's action distribution for selected training iterations in Fig.~\ref{fig:rl-experiments-actions}.
For iteration 0, this is the action distribution of the original SL models.
It can be seen that the model gradually shifts from an initially active policy with few idle actions to a policy that predominantly idles.
While the results show that this is a successful strategy against \agentsimplec, the passive behavior will fail against better opponents.

The decreasing win rate of \agentmctstsl is consistent with these findings.
While \agentmctsosl assumes the opponents to behave like \agentsimplec, \agentmctstsl uses its own policy to expand the moves of the closest opponent.
By selecting idle with a higher probability, the opponent model diverges more and more from the actual opponent playing behavior and the win rate of this approach decreases.

\subsection{Learned Opponent Models}
In the previous sections, we investigated our search methods in combination with supervised and reinforcement learning.
However, we still used the heuristic \agentsimplec as the opponent model during planning.
As \agentsimplec is clearly suboptimal, this may lead to problems when applying the agent against opponents with strategies that differ significantly.
Consequently, we explore the usage of \agentrawnet opponent models within this section.
As the combination of our model learned from demonstrations with our simple action filter is apparently better than \agentsimplec, agents using \agentrawnet opponent models should be capable of adapting to better players.

Tab.~\ref{tab:learned-opponent-models} shows the results of our search approaches for \agentsimplec and \agentrawnet opponent models against \agentsimplec opponents.
The search uses 1000 simulations to be comparable to the results from Fig.~\ref{fig:search-experiments}.
We focus on \agentmctso in the \ac{rl} setting due to the weak performance of \agentmctst and show the results for the models initialized from zero (RL) and the ones initialized with SL and refined with RL (\sltorl).
For both, we use the models after $15$ training iterations due to the peak in Fig.~\ref{fig:rl-experiments}.
All experiments were performed for each of the five respective models and we report the mean results.
Ties include episodes that are not done.

For the SL models, we can see that the win, tie rates and environment steps are nearly unaffected when using \agentrawnet instead of \agentsimplec for both of our approaches.
However, it is noticeable that the search depth increases slightly and the search time increases drastically.
We hypothesize that the increase in search depth is caused by the better opponent behavior, i.e.~episodes within the search do not end as quickly as \agentrawnet is a stronger opponent.
However, as the actual opponents are still \agentsimplec, this is not reflected in the real environment, as visible in similar numbers of environment steps.
The increase in search time can be explained by our prototypical implementation of the \agentrawnet opponent models.
While the model inference for \agentmctso and \agentmctst is executed in batches, we currently use batch size one for \agentrawnet opponent models within the search.
This drastically increases the time to evaluate the opponent models per step.
With an halved search depth in \agentmctst, the search time also decreases greatly as there are fewer inference calls of the opponent model.

For the \sltorl and RL models, we notice that the win rate slightly decreases when using \agentrawnet opponent models.
However, the \sltorl model yields higher win rates than \agentmctso with the SL model and is comparable to \agentmctst with the SL model.
The win rate of the RL model without training on demonstrations is similar to the initial SL model for \agentmctso.
For the \sltorl and RL models, there is high increase in environment steps compared to the SL models.
This indicates that the agents play more passively.

We conclude that in most cases, the \agentrawnet opponent model has a neglectable effect on the win rate against opponents that were seen in training.
With a more efficient implementation, it could be a good alternative to hand-crafted heuristics.
The best agents are \agentmctst with the SL model and \agentmctso with the \sltorl model.

\subsection{Evaluation in the Official Environment}

Finally, we evaluate our approach in the official Python environment \cite{resnick2018pommerman} to investigate whether our previous results transfer to the Python environment and the \agentsimplep opponent.
The starting positions are randomized to reduce their influence on the results.
As the previous section showed a high search time when using \agentrawnet, we restrict the number of simulations to 250 in this case to stay below the official time constraint of 100 ms.

\begin{table}[t]
    \small
    \centering
    \caption{Win rate and tie rate of our approaches against \agentsimplep opponents in the official environment for 100 games. We use 1000 simulations for \agentsimplec and 250 simulations for \agentrawnet.
    All results are averaged over five models.}
    \begin{tabular}{cccccc}
        \toprule
        \textbf{Model} & \textbf{Method} & \textbf{Opponent Model} & \textbf{Win Rate} & \textbf{Tie Rate}\\ \midrule
        \multirow{4}{*}{SL} & \multirow{2}{*}{\agentmctso} & \agentsimplec & $\mathbf{0.78 \pm 0.04}$ & $0.12 \pm 0.02$\\
        & & \agentrawnet & $\mathbf{0.67 \pm 0.06}$ & $0.25 \pm 0.04$\\
        &\multirow{2}{*}{\agentmctst} & \agentsimplec & $0.66 \pm 0.05$ & $0.32 \pm 0.05$\\   
        & & \agentrawnet & $0.61 \pm 0.06$ & $0.34 \pm 0.06$\\
        \midrule
        \multirow{2}{*}{\sltorl} & \multirow{2}{*}{\agentmctso} & \agentsimplec & $0.65 \pm 0.09$ & $0.30 \pm 0.09$\\
        & & \agentrawnet & $0.63 \pm 0.05$ & $0.32 \pm 0.04$\\
        \multirow{2}{*}{RL} & \multirow{2}{*}{\agentmctso} & \agentsimplec & $\mathbf{0.74 \pm 0.03}$ & $0.22 \pm 0.03$\\
        & & \agentrawnet & $\mathbf{0.72 \pm 0.04}$ & $0.23 \pm 0.03$\\
        \bottomrule
    \end{tabular}
    \label{tab:python-eval}
\end{table}

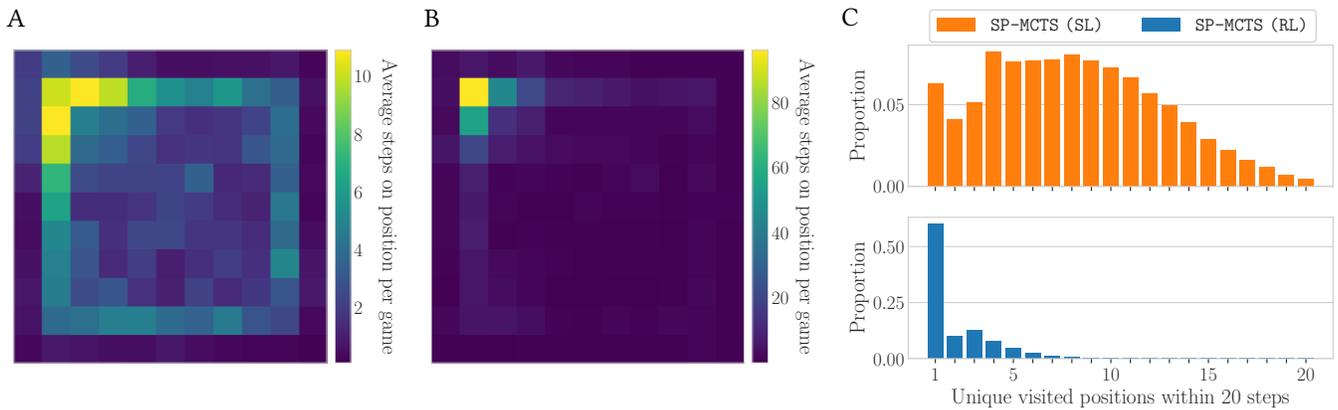
\begin{figure*}[t]
    \centering
    \newcommand{\movementPlotHeight}{4.4cm}
    \newcommand{\thirdPlotHeight}{5.5cm}
    \begin{tikzpicture}
      \node[inner sep=0pt] (a) {\includegraphics[height=\movementPlotHeight,trim={0 0 0 0cm},clip]{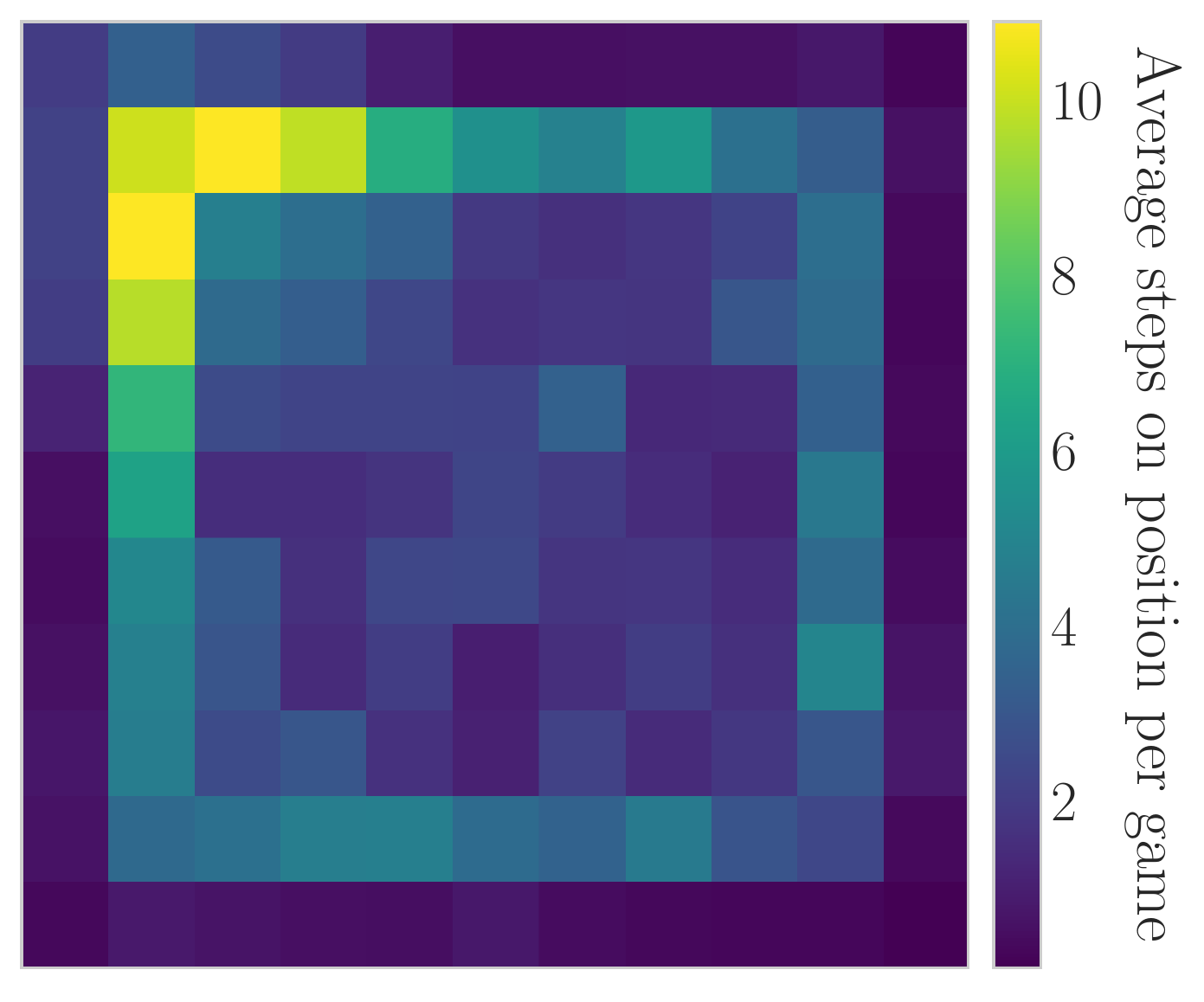}};
      \node[above right, inner sep=0pt] at ([shift={(0,0.2cm)}]a.north west) {\large A};
      \node[inner sep=0pt] (b) [right= 0.25cm of a] {\includegraphics[height=\movementPlotHeight,trim={0 0 0 0},clip]{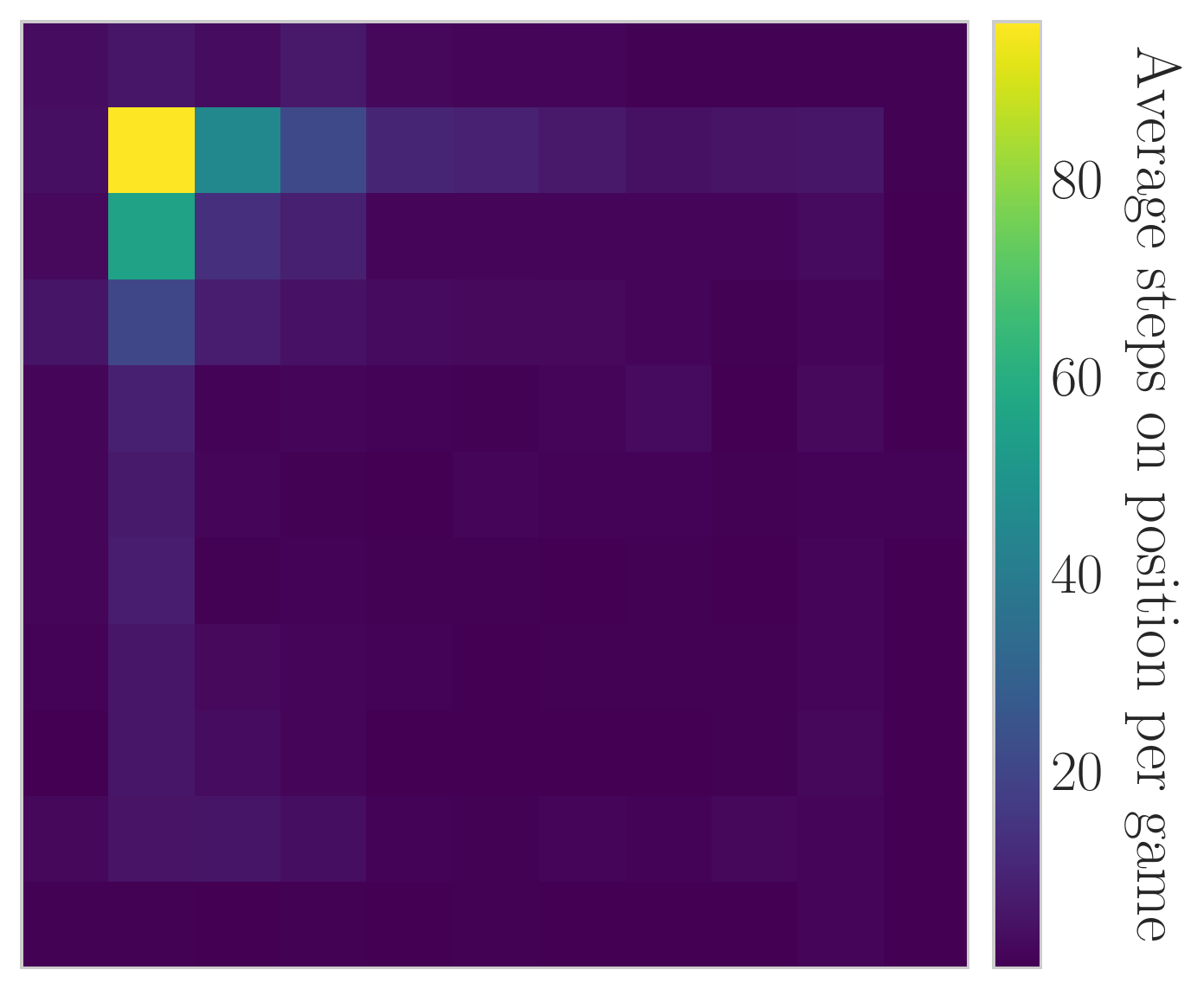}};
      \node[above right, inner sep=0pt] at ([shift={(0,0.2cm)}]b.north west) {\large B};
      \node[inner sep=0pt] (c) [right= 0.25cm of b] {\includegraphics[height=\thirdPlotHeight,trim={0 0 0 0},clip]{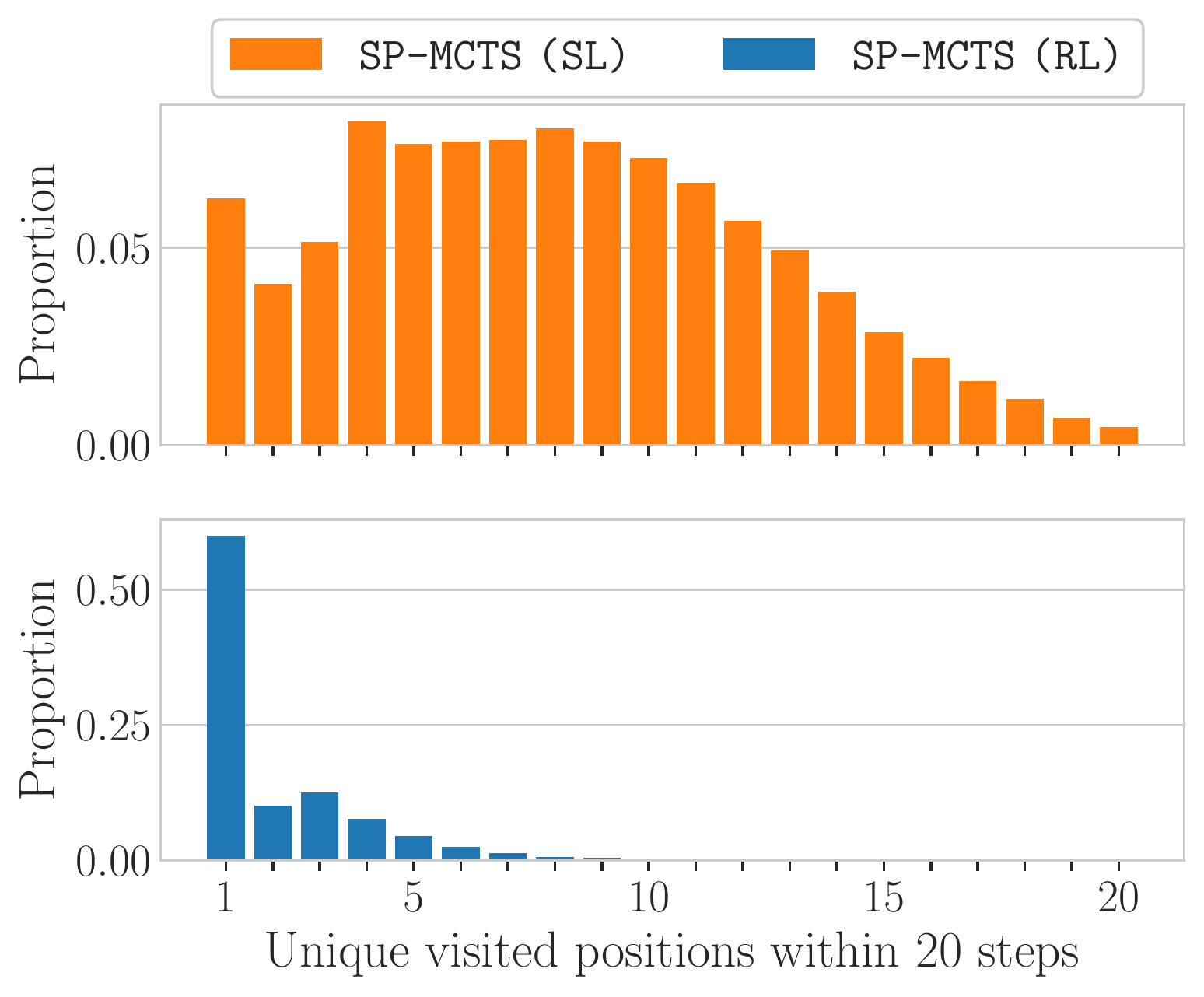}};
      \node[above right, inner sep=0pt] at ([shift={(0,-0.33cm)}]c.north west) {\large C};
    \end{tikzpicture}
    \caption{Visited board positions of the \agentmctso SL (A) and RL agents (B) against \agentsimplep in the official environment. Subfigure (C) complements this with the number of unique positions visited within $20$ steps. While the \agentmctso SL agents show active movement behavior, the agents trained with RL are very passive. The results are averaged over 5 models with 100 games each.}
    \Description{Three subfigures labeled A, B and C. A and B show heatmaps of the visited board positions of the SP-MCTS SL and RL agents against SimplePy in the official environment. Subfigure C complements this with bar plots for both agents that illustrate the number of unique positions visited within 20 steps. While the SP-MCTS SL agents show active movement behavior, the agents trained with RL are very passive. This is indicated by very low color intensities except for the starting area in B and by fewer unique visited positions in C. The results are averaged over 5 models with 100 games each.}
    \label{fig:behavior-python}
\end{figure*}

The results against \agentsimplep agents in the \ac{ffa} mode are shown in Tab.~\ref{tab:python-eval}.
In contrast to our previous evaluation  against \agentsimplec in Tab.~\ref{tab:learned-opponent-models}, \agentmctso outperforms \agentmctst by a noticeable margin for the SL models.
While the results of \agentmctso are similar to our previous evaluation, the win rate of \agentmctst strongly decreases.
A potential reason for that could be the difference between the actual opponent behavior and the one considered during planning. 
The search might assume that opponents play too well, resulting in an overly defensive play style.
This is also indicated by the high reduction in the win rate when using the \agentrawnet opponent model in \agentmctso with the SL models.
Another indicator for the defensive playing style is the high tie rate and the high similarity of the results to the \sltorl models.
Interestingly, the RL models outperform the \sltorl models in this setup.
It could be that they generalize better against other opponents because they were not trained on demonstrations.
We notice that the tie rates of all agents are very high, especially for the agents with lower win rates.
Most of these ties are caused by episodes that do not terminate within the environment's limit of 800 steps.
In turn, the average number of environment steps per episode also increases greatly to up to $490 \pm 230$ steps for \agentmctso with the RL model.
We omit the steps in the table, their overall trend for the individual models and methods is similar to the previous results. 
Despite the reduced win rates, our agents still loose very few games due to a defensive play style.

Recent related work with learning-based \ac{mcts} and reward shaping reports a win rate of around $0.7$ against \agentsimplep opponents in the \ac{ffa} mode \cite{yang21MCTSPommerman}.
Our approaches reach competitive win rates without reward shaping.
Additionally, the RL agent was trained from scratch and did not use learning from demonstrations.

We show further details regarding the movement behavior of the \agentmctso SL and RL agents with \agentsimplec opponent models in Fig.~\ref{fig:behavior-python}.
Subfigures (A) and (B) show the average steps on the individual board positions per game.
For the visualizations, we rotated the board according to the agents' starting positions such that they always start at the upper left corner.
This allows us to see how much the agents explore the map, irrespective of their starting position.
In Fig.~\ref{fig:behavior-python} (A), we can see that the \agentmctso SL agent actively explores the map while avoiding the border, except for the tiles close to its starting position.
This is reasonable, as agents at the border have fewer options for evasion.
The noticeable ring across the map at distance one to the border is due to the randomization of the map.
Only destructible objects and passages are placed at these positions, ensuring that the agents can reach each other.
In Fig.~\ref{fig:behavior-python} (B) and (C), we can see that the \agentmctso RL agents stay very close to their starting positions and rarely move across the map, confirming that they develop a very defensive playing style.

\section{Discussion}
Next, we discuss the insights and limitations of our approach.
One major insight is that in Pommerman, focusing on the win rate alone is not enough.
While a high win rate is an indicator for subjectively good agent behavior, further analysis of the behavior of the agents is required to assess the quality of their policies.

We have shown that no custom reward shaping is necessary to significantly improve agent models with our search methods.
Our approach \agentmctso can reach proficient level of play in the \ac{ffa} environment and even compensate for a bad model.
\agentmctst can outperform \agentmctso with a higher number of simulations, but only when a good model is given.
Our evaluation with \agentsimplep opponents suggest that there might not be a single opponent model that allows the agents to perform well in all cases.
If the real opponents show suboptimal behavior that is not captured by the opponent model, our approaches become overly defensive.
One limitation of our experiments is that we only considered deterministic opponent models and models trained on \agentsimplec agents.
Instead, one could collect samples from the best available agents and train models to imitate their behavior.
We expect that combining these models with our search approaches could further increase their playing strength.
Stochastic opponent models could be considered by expanding multiple different opponent actions into individual nodes or merging different opponent trajectories into a single node with the expected or worst-case behavior. This would greatly increase the sample complexity, but could reduce wrongful exploitation when facing different opponents and make the search more applicable to realistic scenarios.
Another direction for future work lies in the way opponents are selected by \agentmctst.
In our case, we always expand the actions of the closest opponent.
Extensions could predict the most dangerous opponent or selectively expand opponents based on a computational budged.
Finally, we think that combining more computationally expensive search methods  \cite{esser2013BRSplus, baier2020opponentPruningParanoid} with learned opponent models poses an interesting direction for future research.

While \ac{rl} improved the win rate of some configurations, the resulting behavior did subjectively deteriorate.
In particular, the agents trained via \ac{rl} were predominantly passive.
Providing a fine-grained reward signal might be necessary to learn the desired behavior, at the cost of introducing additional bias.
Another interesting aspect of \ac{rl} would be to investigate pure self-play, e.g.~having four learning \agentmctso agents play against each other.
Preliminary experiments not discussed in this paper suggest that self-play agents also develop a passive playing style, further strengthening the need for intermediate rewards when using \ac{rl}.
A different approach to stabilize training in a self-play setting could be to anticipate the learning of other agents in the environment \cite{foerster2018learning}.
One problem that emerges without self-play is that high win rates lead to unbalanced data sets.
We experimented with resampling techniques to draw samples for each value and action target with equal probability, but this did not lead to noticeable improvements.
To avoid overfitting on specific opponents, population-based approaches with different opponents would also be worth investigating.

Lastly, we focused on the \ac{ffa} mode.
Extensions of our approach to the team and radio mode would be interesting. 
Initial results show that our agent performs well against a team of \agentsimplep opponents, but struggles against docker agents from previous challenges.
To deal with the partial observability in form of the now limited view, agent models leveraging recurrent neural networks and learned communication can be explored.
We think that including bomb kicks in the demonstration data set, e.g. by considering samples from different agents, could greatly help our agents to react appropriately when facing these opponents.
It would also be interesting to combine our approaches with learned environment models, especially to avoid hand-crafting environment dynamics that can handle the limited view in the team and radio modes.

\section{Related Work}
The combination of \ac{rl} and tree search has been extensively studied in the domain of games, in particular board games.
One breakthrough in this field is the work by \citet{anthony2017thinking}, which applied tree search and reinforcement learning to learn the game of Hex.
This approach was later followed and popularized in the AlphaZero \citep{silver2018general} algorithm by learning the games Go, Shogi and Chess from zero knowledge.
AlphaZero has then been re-implemented and extended for Go \cite{wu2019accelerating}
and multiple chess variants \cite{czech2021improving}.
Later works in the form of MuZero \cite{schrittwieser2020mastering} emphasized the environment model by learning a model that is used for planning rather than relying on the actual environment itself.

Planning approaches using tree search have also been successfully applied in multiplayer games with more than two players \cite{luckhart1986maxn, nijssen2013phdMultiplayerMCTS}.
However, they can suffer from a shallow search depth \cite{baier2020guidingMultiplayerMCTS} in practice due to the high combinatorial complexity.
Many approaches increase the search depth by reducing the time spent to simulate opponents and expectedly suboptimal moves \cite{sturtevant2000paranoidSearch, schadd2011bestReplySearch, baier2020opponentPruningParanoid}.
In addition to simplifying the search tree, integrating domain knowledge in the form of value heuristics into \ac{mcts} has shown to greatly improve performance in multiplayer games \cite{szita2009monte, nijssen2011playout}.
\citet{petosa2019multiplayer} extend the idea of AlphaZero and apply learned value estimation in multiplayer games, but iterate over all players during search.
On contrast, \citet{ozair21quantizedModels} consider opponents to be part of the environment's dynamics and learn a latent variable to sample state transitions for \ac{mcts} that include the opponent's moves.
While they only considered games with up to two players in their evaluation, the idea should be generally applicable.

Previous work in Pommerman \cite{resnick2018pommerman} ranges from learning- and planning-based approaches to the combination of both.
Due to the sparse reward and long time horizon, approaches leveraging model-free \ac{rl} struggle to beat simple heuristics without further modifications of the environment or training procedure \cite{gao2019PommermanExploration}.
\citet{resnick18Backplay} suggest to start training close to terminal states,
\citet{peng18COMBAT} use pathfinding-based actions instead of direct movement and
\citet{gao19SkyNet} employ reward shaping, action filtering and curriculum learning.
However, agents using model-free \ac{rl} have shown inferior performance compared to planning-based approaches.
Their main limitations are the time constraints for decision making and the high branching factor.
The winners of the NeurIPS 2018 competition combine planning with deterministic and pessimistic rollouts to increase the search depth \cite{osogami2019real}.
Rollouts with more than ten steps allow the agents to account for explosions of recently placed bombs during planning.
The second-placed agent uses minimax search with an average search depth of only two steps \cite{NeurIPS18Competition},
an extension of this agent won the subsequent competition held at NeurIPS 2019.
Learning and planning can also be combined. 
For example, \citet{kartal19PommerMCTS} use model-free \ac{rl} but initialize their agent with imitation learning on samples generated by shallow \ac{mcts} with random agents.
\citeauthor{yang21MCTSPommerman} combine \acs{mcts} with a learned model, they initialize their agent with imitation learning and employ reward shaping and sophisticated action filtering heuristics during search \cite{yang21MCTSPommerman}.

We observe that, although they reach proficient levels of play, recent planning-based approaches either suffer from a shallow search depth or introduce high bias through search heuristics and reward shaping.
In this paper, we explored the feasibility of learning-based \ac{mcts} in the Pommerman environment with opponent models, given only an environment model, a sparse reward signal at the end of the episode, and demonstrations from other agents.

\section{Conclusion}
With this work, we proposed two methods based on \ac{mcts} that make use of deterministic opponent models and reduce competitive multiplayer games to single- and two-player games.
This greatly reduces the complexity of the search space and makes \ac{mcts} applicable to complex environments with time constraints.
We evaluated our approach in the game Pommerman without custom reward shaping.
We found that both methods load to high improvements in terms of win rate against baseline agent heuristics, both when using an uninitialized model and a model trained on demonstrations.
Our two-player search outperforms the single-player search, but requires more simulations and a good initial model.
While the application of \ac{rl} based on the samples generated by the search leads to improved win rates in most cases, we found that the agents develop a passive playing style.
We think that intermediate rewards might be necessary to learn a more active policy in a \ac{rl} setup.

Future work could investigate how our approaches perform if demonstrations from better agents are used to train the initial models.
To further explore the \ac{rl} setup, the next step would be to integrate reward shaping.
It would also be interesting to expand upon the opponent selection in our two-player search, e.g.~by predicting the most dangerous opponent in each step.  
Another promising direction would be to consider \ac{mcts} with stochastic opponent models.
To extend our approach to the team and radio modes, the combination of \ac{mcts} with recurrent neural networks and learned communication between agents could be explored further.

%% file: acknowledgments.tex
The authors thank Jonas Ringsdorf for assisting with the implementation and evaluation, Adrian Alic for the development of pomcpp, and Quentin Delfosse as well as Jannis Blüml for valuable feedback. This work benefited from the Hessian Ministry of Science and the Arts (HMWK) project ‘The Third Wave of AI' and has been co-funded by the German Research Foundation (DFG) in the Collaborative Research Center (CRC) 1053 MAKI.